\documentclass[letterpaper, 10 pt, conference]{ieeeconf}  % Comment this line out if you need a4paper

\IEEEoverridecommandlockouts                              % This command is only needed if 
\usepackage[cmex10]{amsmath}
\usepackage[pdftex]{graphicx}
\usepackage{graphicx}
\usepackage{float}
\usepackage{array}
\usepackage{amsmath}
\usepackage{cases}
\usepackage{verbatim}
\usepackage{lipsum}
\usepackage{siunitx}
\usepackage[]{cite}
\usepackage[caption=false,font=footnotesize]{subfig}
\usepackage{wrapfig}
\usepackage{dblfloatfix}
\usepackage[symbol]{footmisc}

\title{\LARGE \bf
Sensorimotor-inspired Tactile Feedback and Control Improve Consistency of Prosthesis Manipulation in the Absence of Direct Vision
}

\author{Neha Thomas$^{1}$, Farimah Fazlollahi$^{2}$, Jeremy D. Brown$^{3}$, and Katherine J. Kuchenbecker$^{2}$% <-this % stops a space
\thanks{$^{1}$Neha Thomas is with the Department of Biomedical Engineering, Johns Hopkins School of Medicine, and the Haptic Intelligence Department, Max Planck Institute for Intelligent Systems, {\tt\small neha.thomas@jhmi.edu}}% <-this % stops a space
 
\thanks{$^{2}$Farimah Fazlollahi and Katherine J. Kuchenbecker are with the Haptic Intelligence Department, Max Planck Institute for Intelligent Systems, {\tt\small fazlollahi@is.mpg.de} and {\tt\small kjk@is.mpg.de}}
        
\thanks{$^{3}$Jeremy D. Brown is with the Department of Mechanical Engineering, Johns Hopkins University, {\tt\small jdelainebrown@jhu.edu}}%
}

\begin{document}

\maketitle
\thispagestyle{empty}
\pagestyle{empty}

%%%%%%%%%%%%%%%%%%%%%%%%%%%%%%%%%%%%%%%%%%%%%%%%%%%%%%%%%%%%%%%%%%%%%%%%%%%%%%%%
\begin{abstract}
The lack of haptically aware upper-limb prostheses forces amputees to rely largely on visual cues to complete activities of daily living. 
% However, this approach is limiting because many situations do not allow for constant visual attention. 
In contrast, able-bodied individuals inherently rely on conscious haptic perception and automatic tactile reflexes to govern volitional actions in situations that do not allow for constant visual attention. We therefore propose
% In contrast, able-bodied individuals are easily able to substitute haptic for visual guidance in dexterous tasks. This guidance comes in the form of conscious haptic perception and autonomous tactile reflexes, which are managed by the sensorimotor system. 
a myoelectric prosthesis system that reflects these concepts to aid manipulation performance without direct vision. To implement this design, we built two fabric-based tactile sensors that measure contact location along the palmar and dorsal sides of the prosthetic fingers and grasp pressure at the tip of the prosthetic thumb. Inspired by the natural sensorimotor system, we use the measurements from these sensors to provide vibrotactile feedback of contact location and implement a tactile grasp controller that uses automatic reflexes to prevent over-grasping and object slip. We compare this system to a standard myoelectric prosthesis in a challenging reach-to-pick-and-place task conducted without direct vision; 17 able-bodied adults took part in this single-session between-subjects study.  Participants in the tactile group achieved more consistent high performance compared to participants in the standard group. These results indicate that the addition of contact-location feedback and reflex control increases the consistency with which objects can be grasped and moved without direct vision in upper-limb prosthetics.

% Users of standard myoelectric upper-limb prostheses must rely largely on visual cues to complete activities of daily living. However, this approach is limiting because many situations do not allow for constant visual attention.  Tactile sensing can help: various forms of haptic feedback and autonomous grasp controllers based on tactile cues have been suggested to reduce reliance on vision and improve object manipulation performance. This paper thus describes two fabric-based tactile sensors that measure contact location along the exterior and interior of prosthetic fingers and grasp pressure at the tip of the prosthetic thumb. Inspired by natural sensorimotor systems, we use the measurements from these sensors to provide vibrotactile feedback of contact location and implement a tactile grasp controller that uses automatic reflexes to prevent over-grasping and object slip. We compare this system to a standard myoelectric prosthesis in a challenging reach-to-pick-and-place task conducted without direct vision; 17 able-bodied adults took part in this single-session between-subjects study.  Participants in the tactile group achieved more consistent high performance than participants in the standard group. These results indicate that the addition of contact-location feedback and reflex control increases the consistency with which objects can be grasped and moved without direct vision in upper-limb prosthetics. 

\end{abstract}

%%%%%%%%%%%%%%%%%%%%%%%%%%%%%%%%%%%%%%%%%%%%%%%%%%%%%%%%%%%%%%%%%%%%%%%%%%%%%%%%
\section{Introduction}

Sensorimotor control is classically divided into two domains: volitional control and reflexive control. Volitional movement results from high-level cognitive processing, while reflexes begin with sensory perception \cite{Schwartz2016}. The process of picking up and relocating an object, such as a pen, involves both type of movements -- a volitional move to grasp followed by fine, quick grip force adjustments based on sensed properties like weight and friction \cite{Johansson2009}. 
 
With direct visual observation, reaching for and grasping everyday objects is trivial. When visual attention must be directed elsewhere, this task becomes less trivial but is still relatively easy for able-bodied individuals. In memory-guided reach-to-grasp tasks without visual feedback, hand posture during movement updates similarly to that with visual guidance, with the hand conforming to the contours of the object upon contact \cite{Santello2002}. Tactile cues then help determine motor coordination in the fingers in order to grasp and lift the object \cite{Johansson1992}.
 
Searching for a coin in a pocket or picking up a pen while staring at a screen are tasks that able-bodied individuals accomplish predominantly through haptic sensations that drive normal sensorimotor control. Upper-limb amputees using a myoelectric prosthetic, however, do not have this luxury. Because standard commercial prosthetics lack the sensory and reflexive properties of the intact human limb, amputees must rely extensively on visual feedback. Although vision can compensate for some of the missing haptic information \cite{Sensinger2020}, there are situations where vision cannot be so heavily depended on or is undesirable, such as when lighting conditions are poor, when objects are occluded, or when multitasking -- for example, cooking while watching a tutorial. 
% one is watching a child, a pot of boiling water, or a virtual meeting.%multi-tasking, as vision is known to be cognitively demanding \cite{...}.
Indeed, 438 users of transradial electric-powered prostheses ranked ``required less visual attention to perform functions'' as their third-highest priority on average out of 17 options for system improvement \cite{Atkins96-JPO-Priorities}. 
 
\begin{figure}[t]
		\centering
		\includegraphics[width=\columnwidth]{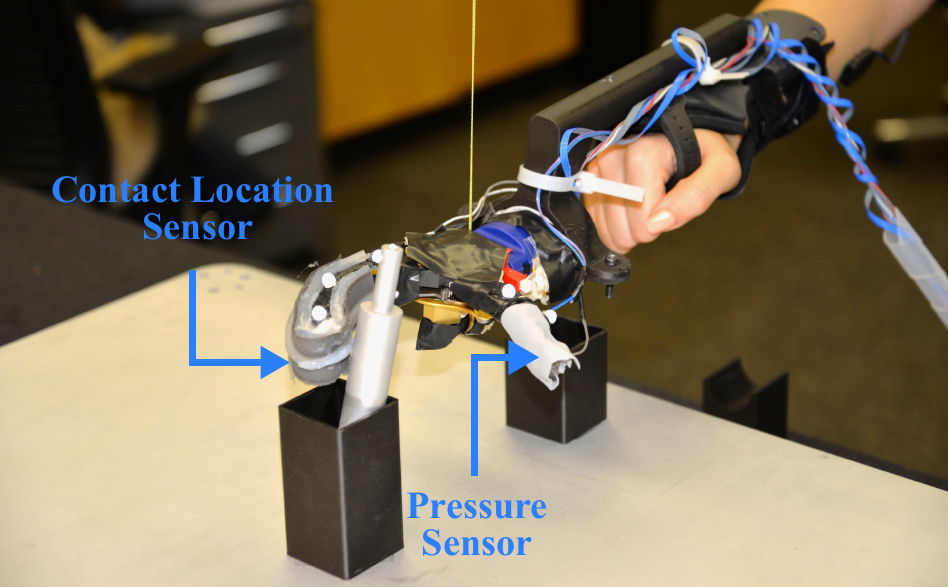}
		\vspace{-1.75em}
		\caption{A myoelectric Ottobock hand fitted with custom tactile sensors being used by an able-bodied individual to complete a difficult everyday manipulation task: picking up a thin cylindrical object without directly looking at the hand or object. Prosthesis users are eager to manipulate objects without direct vision because such situations frequently arise, as when one is visually attending to a conversation partner, another task, or a screen. The sense of touch can naturally fill this perceptual gap.}
		\label{fig:Hand}
		\vspace{-1.25em}
\end{figure}

One well-studied strategy to reduce reliance on vision during prosthesis use is to provide referred haptic feedback of pertinent grasp information \cite{Sensinger2020}. In the context of able-bodied individuals grasping without vision, tactile cues like contact detection and localization are important for updating hand posture to grasp an object correctly. In this way, these discrete contact events aid volitional control. This paradigm aligns with discrete event-driven sensory control (DESC) theory, which describes milestone phases of grasp and lift as discrete events marked by bursts of activity in tactile afferents \cite{Johansson1993DESC}. Previous research applying DESC theory to a myoelectric prosthesis operated by transradial amputees used vibrotactile stimulation to provide feedback of object contact and release. This system improved user performance in a virtual egg task \cite{Clemente2016b}. Similarly, when vibrotactile feedback of discrete force increments was added to a myoelectric prosthesis, both grasping errors and completion time reduced in a box-and-blocks task performed with reduced visual feedback \cite{Raveh2018a}.
 
Another potential way to reduce reliance on vision is to implement autonomous grasp controllers that respond to tactile events such as contact, slip, or sensed cues such as grasp force. These controllers mimic the natural reflex pathways of the intact limb without directly involving the user. Previous research has shown fewer object breaks and faster task completion during object manipulation with autonomous grip control \cite{Osborn2016}. In addition, such controllers have already been implemented into some commercial prosthetic hands, like the Ottobock SensorHand Speed \cite{Ottobock}.

To date, however, there has been no prior research on combining haptic feedback with autonomous grasp control into a cohesive system that mimics the biological principles of volitional and reflexive control. Likewise, it is not clear how manipulation performance with such a prosthesis would compare to a standard commercial prosthesis. 
% To date, however, there has been no work comparing a standard clinical prosthesis with a prosthesis that provides haptic feedback and autonomous grasp control, 
In this paper, we describe the development of a bioinspired system for prosthesis control that utilizes a custom-built pressure sensor and a novel contact-location sensor on the fingers of a commercial prosthetic hand (Fig.~\ref{fig:Hand}). Together the signals from these sensors enable both volitional human-in-the-loop control through contact-location haptic feedback and reflexive autonomous grasp control through closed-loop tactile sensing. We conduct a user study featuring a reach-to-pick-and-place task to compare the performance of this bioinspired prosthesis to a standard commercial prosthesis. Our investigation focuses on the utility of this system in a dexterous task without direct visual observation, a common scenario that is similar to picking up and drinking from a cup of coffee while looking at a computer screen. 
% This common is diverted - similar to how able-bodied individuals are able to fixate on their screen during a videocall while grabbing a pen.

\begin{figure}[t]
		\centering
		\vspace{1em}
		\includegraphics[width=\columnwidth]{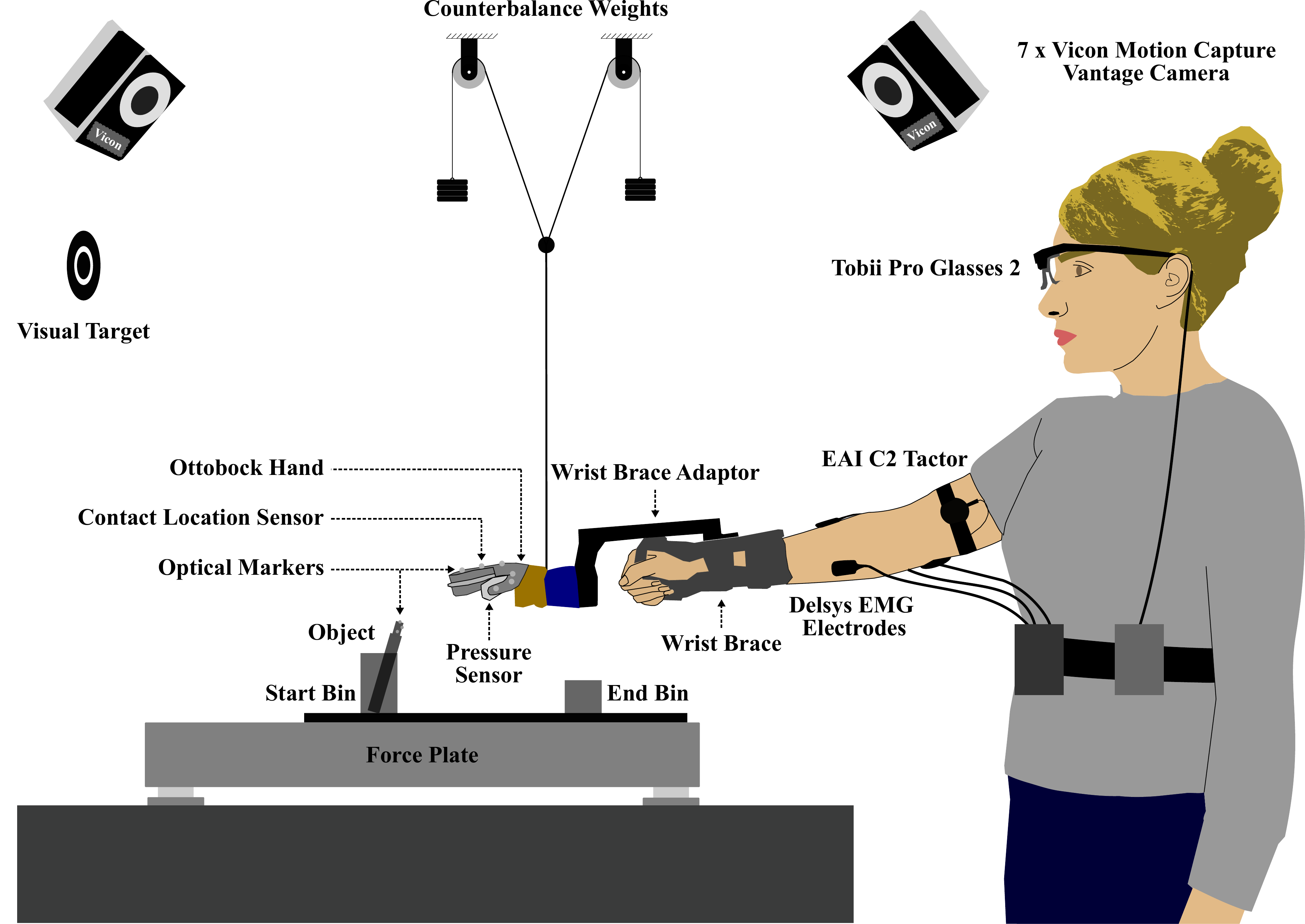}
		\vspace{-1.5em}
		\caption{The experimental task involves using a myoelectrically controlled prosthesis to find, grasp, pick up and move a cylindrical object to another bin. To simulate situations that lack direct vision, the participant is required to focus their gaze on a target on the opposing wall throughout the task.}
		\label{fig:setupNoDAQ}
		\vspace{-1em}
	\end{figure}

\section{Methods}

\subsection{Participants}
We investigated the ability of 17 participants (12 male, age 30.7 $\pm$ 4.1 years) to perform a reach-to-pick-and-place task using a myoelectric prosthesis in a between-subjects study with two conditions. The two groups were balanced for gender and handedness; the two left-handed participants and one ambidextrous participant all did the study with the right hand. We chose a between-subjects design to avoid transfer of skill between conditions and to better simulate deployment of a particular prosthesis with a user. All participants were consented according to a protocol approved by the Ethics Council of the Max Planck Society under the Haptic Intelligence Department’s framework agreement as protocol number F005C. The experiment lasted one hour, and participants not employed by the Max Planck Society received 8 euros as compensation. 

\subsection{Experimental Apparatus}
\label{Sec:apparatus}
As shown in Fig.~\ref{fig:Hand}, the prosthesis consists of a one-degree-of-freedom (1-DoF) OttoBock SensorHand Speed myoelectric prosthetic hand attached to a wrist brace through a 3D-printed adaptor so that it can be used by able-bodied individuals. The weight of the hand (approx. 500\,g) was partially offset by a counterweight system (400\,g) to better replicate the loading situation for a transradial amputee who would wear the prosthesis in place of their amputated hand and wrist. The prosthesis fingers were fitted with two custom-made piezoresistive fabric sensors that measure pressure and contact location, as detailed in Section~\ref{Sec:Sensors}. The speed of the hand's grasping DoF is controlled proportionally through surface electromyographic (sEMG) signals from the user's wrist flexor and extensor muscles. sEMG was acquired through an 8-channel Delsys Bagnoli EMG system. Vibrotactile feedback was provided with a C2 tactor (Engineering Acoustics, Inc.) mounted above the biceps muscle and driven by a linear current amplifier. Data acquisition and control was done at 1000\,Hz with MATLAB/Simulink (2019a) and an NI myRIO running on QUARC real-time software (2020 v4.0.3032). The sensor placement and overall setup are shown in Fig.~\ref{fig:setupNoDAQ}.

Motion capture was implemented using a seven-camera Vicon Vantage system with Vicon Nexus software (v2.11.0). Optical markers were placed on the prosthesis as well as the task object. Force data was acquired with a custom-built force plate consisting of four ATI Mini40 force sensors capable of measuring up to 480\,N in the vertical direction. Force signals were acquired with an NI 6255 DAQ.
Tobii Pro Glasses 2 were used to track the user's gaze, which was recorded and processed via iMotions software (v8.2). 
All data streams were synchronized using a mechanical switch to denote the start and end of a study session.

% \begin{figure*}[t]
% 		\centering
% 		\includegraphics[width=\textwidth]{Figures/setup.pdf}
% 		\caption{}
% 		\label{fig:setup}
% 	\end{figure*}

\subsubsection{sEMG Calibration \& Myoelectric Control}
\label{Sec:calibration}
To calibrate the sEMG for the wrist flexor and extensor muscles, participants first donned the prosthesis device. Next, they were asked to lift their arm and hold still for 5 seconds for a baseline sEMG measurement. Finally, they were asked to perform wrist flexion and extension at maximum voluntary contraction (MVC) for 5 seconds each. The offset was calculated for each of the sEMG signals using the baseline measurement. 50\%  of the average flexion was used as the upper threshold for flexion. The lower threshold for flexion was set to the average contraction of the flexor muscle during the extension MVC or to 5\% of the flexion MVC, whichever was higher. The same was done to calculate the upper and lower thresholds for the extension signal.

The offset was added to each of the sEMG signals. Then, using the respective lower and upper thresholds, each of the signals was normalized between the minimum and maximum voltages for the Ottobock hand's motor. The control law for closing the hand was
\begin{equation}
u_{c} = \begin{cases}
    S_{f} &,\ S_{f} - S_{x} > 0\\
    0 &,\ \text {otherwise}\\
  \end{cases}
  \label{eqn:closing}
\end{equation}
\noindent where $S_{f}$ is the normalized flexor signal, and $S_{x}$ is the normalized extensor signal. Similarly, the control law for opening the hand was
\begin{equation}
u_{o} = \begin{cases}
    S_{x} &,\ S_{x} - S_{f} > 0\\
    0 &,\ \text {otherwise}\\
  \end{cases}
  \label{eqn:opening}
\end{equation}
\noindent where $S_{f}$ and $S_{x}$ are the same as in (\ref{eqn:closing}).

\subsubsection{Sensors}
\label{Sec:Sensors}
Both sensors used on the prosthesis were custom designed and constructed; they are shown in Fig. \ref{fig:Hand}.

\paragraph{Pressure Sensor}
The pressure sensor was placed on the thumb of the prosthesis to measure grasp force. It consists of three layers of fabric: the top and bottom layers are conductive fabric, and the middle layer is piezoresistive fabric. The sensor functions as a variable resistor whose resistance reduces with pressure; the design is based on work by Osborn {\it et al.} \cite{Osborn2014}. The sensor is implemented in a voltage divider circuit, whereby a voltage is applied across the sensor and a 1\,k$\Omega$ resistor connected in series with it. As pressure on the sensor increases, the measured voltage across the 1\,k$\Omega$ resistor increases. 

\begin{figure}[t]
		\centering
		\includegraphics[width=\columnwidth]{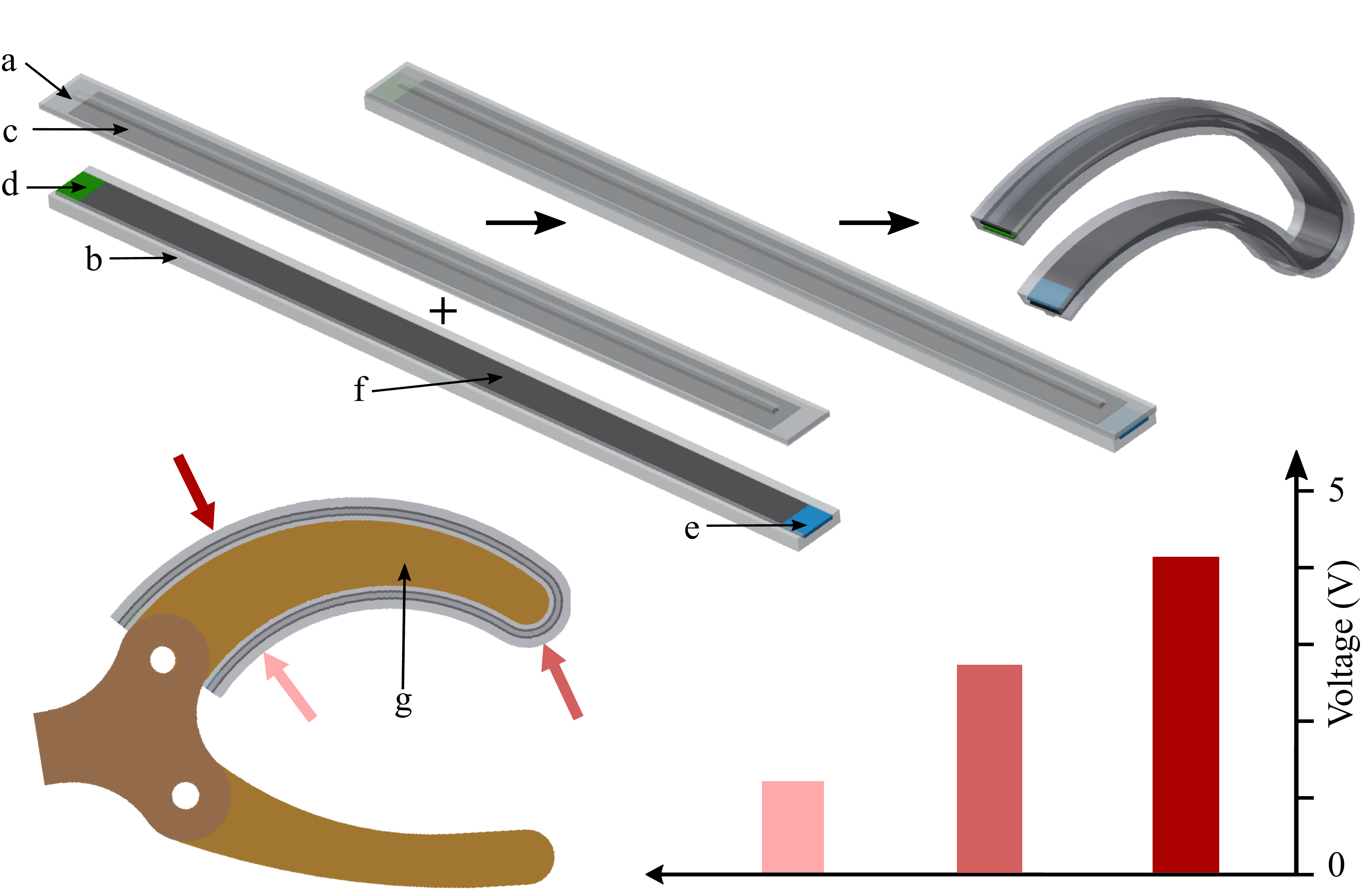}
		\vspace{-1.5em}
		\caption{Contact-location sensor: a) and b) show the silicone frames of the top and bottom layers of the sensor, respectively. Item c) is the long layer of conductive fabric; d) and e) are the conductive fabric electrodes attached to the two ends of f) the lower piezoresistive fabric layer, which has a voltage gradient across its length. The sensor is wrapped around the curved external surface of g) the prosthesis fingers. When something touches the outside of the sensor, the top layer of conductive fabric touches the bottom layer of piezoresistive fabric at the contact point. The voltage measured on the conductive fabric can be used to identify the location point along the length of the sensor. }
		\label{fig:ContactSensor}
		\vspace{-1.25em}
	\end{figure}

\paragraph{Contact-location Sensor}
The contact-location sensor was wrapped around the fingers of the prosthesis and covers both the palmar and dorsal sections. 
The sensor consists of one layer of piezoresistive fabric and one layer of conductive fabric. Both layers are fixed within an outer silicone frame (SmoothOn Ecoflex 30) and separated from each other by a thin air gap. A voltage gradient is created across the length of the piezoresistive layer, so that a distinct voltage is elicited depending on where the conductive layer contacts the piezoresistive layer, as conceptually demonstrated in Fig.~\ref{fig:ContactSensor}.

\subsubsection{Sensorimotor-inspired System}
The bioinspired system includes vibrotactile feedback of contact location, which aims to aid haptically guided volitional control, and autonomous reflex controllers, which react to tactile events.
\paragraph{Vibrotactile feedback of contact location}
\label{Sec:VibroFeedback}
Vibrotactile feedback was provided by a C2 tactor (Engineering Acoustics, Inc.) which was driven according to signals from the contact-location and pressure sensors. The contact-location sensor signals were first normalized between 0 (proximal) and 1 (distal), with respect to the prosthesis finger, regardless of whether the contact was on the palmar or dorsal side of the finger.

Contact on the dorsal aspect of the fingers was mapped to a constant vibration, while contact on the palmar aspect caused a pulsing vibration. Contact near the tip of the fingers elicits a lower intensity vibration than contact at the proximal part of the fingers. The mapping from the normalized sensor signal $X$ to the current input to the C2 tactor $I$ for both dorsal and palmar finger contact is as follows: 
\begin{equation}
I = \begin{cases}
    0.5\,\textrm{A} \cdot \sqrt{1-X} \cdot \sin{(2 \pi \cdot 250\,\textrm{Hz} \cdot t)} &, \, \text{dorsal}\\ 
    E(t) \cdot 0.5\,\textrm{A} \cdot \sqrt{1-X} \cdot \sin{(2 \pi f t)} &, \, \text {palmar}\\
  \end{cases}
  \label{Eqn:VibMapping}
\end{equation}
% (\ref{Eqn:VibMapping}), with separate equations to differentiate contact on the dorsal from the palmar side of the fingers. 
%
%
\noindent where $E(t)$ is an envelope function denoted by $|\sin{(2 \pi \cdot 4.75\,\textrm{Hz} \cdot t)}|$. % and $f$ is a parameter that is determined as shown below
%
%\begin{equation}
%f = \begin{cases}
%    250\,\textrm{Hz} &, \, p < p_{thr}\\ 
%    250\,\textrm{Hz} -  (50\,\textrm{Hz/s}) t \,dt &, \, p \geq p_{thr}, \, t<2\,\textrm{s}\\
%    150\,\textrm{Hz} &, \, p \geq p_{thr}, \, t\geq 2\,\textrm{s}
%
%  \end{cases}
%  \label{Eqn:Frequency}
%\end{equation}
%
%\noindent where $p$ is the pressure sensor signal, $p_{thr}$ is a heuristically determined threshold, and $t$ is the time elapsed after the contact started. 
When an object is grasped and the pressure sensor signal exceeds a heuristically determined threshold $p_{g}$, the frequency of the vibration $f$ linearly decreases from 250\,Hz to 150\,Hz over a period of 2\,s, well within the range of frequencies easily detected by humans~\cite{Morioka2008VibrotactileHeel}.

\paragraph{Reflex Controllers}
\label{Sec:AutoControl}
Both algorithms described here are based on work by Osborn {\it et al.} \cite{Osborn2016}.

\indent \indent \indent 
{\it i) Over-grasp Controller:}
This controller prevents over-grasping by modulating the closing command $u_{c}$ to the motor according to the control law
\begin{equation}
u_{c} = \begin{cases}
      u_{c} \cdot e^{-K \cdot p} &, \, p \geq p_{g}, \  \ \text {palmar}\\ 
    u_{c} &, \, \text {otherwise}\\

  \end{cases}
  \label{Eqn:AntiOverGrasp}
\end{equation}
\noindent where $K$ is the gain, $p$ is the pressure sensor signal, and $p_{g}$ is the pressure threshold for detecting grasp. 

\indent \indent \indent
{\it ii) Anti-slip Controller:}
The pressure sensor signal was used to determine when slip events occurred and to differentiate between fast and slow slip. Fast slips were determined by looking for rapid decreases in the pressure, as follows: 
\begin{equation}
{\text{Slip}}_{f} = \begin{cases}
      1  &, \, \frac{dp}{dt} \le {q}_{fs}\\
     0 &, \, \text {otherwise}\\
   \end{cases}
  \label{Eqn:Fastlip}
\end{equation}

\noindent where $\frac{dp}{dt}$ is the time derivative of the pressure sensor signal and $q_{fs}$ is a negative threshold. When a fast slip occurs, a closing command is sent to the motor at maximum voltage for 60\,ms to prevent the object from falling out of the hand, which is similar to the reaction times for grasp tightening in human response to slip \cite{Cole1988GripObject}.

Slow slips were determined as follows:
\begin{equation}
{\text{Slip}}_{s} = \begin{cases}
      1  &, \, p(t) - p(t-0.5\,\textrm{s}) < {p}_{ss} \\ 
      0 &, \, \text {otherwise}\\
  \end{cases}
  \label{Eqn:SlowSlip}
\end{equation}
\noindent where $p(t)$ is the pressure sensor signal at the current time point (measured in seconds) and $p_{ss}$ is a negative threshold. When a slow slip occurs, a closing command is sent to the motor at maximum voltage for 30\,ms.
Response time from slip detection to grasp activation was less than 1\,ms.

\subsection{Experimental Protocol}

\subsubsection{Experiment Task}
Participants were asked to complete a reach-to-pick-and-place task with an aluminum cylindrical object (12\,cm long, 2\,cm diameter) using the myoelectric prosthesis. The cylinder approximates the size and shape of many objects that are encountered in daily life, such as a whiteboard marker or a screwdriver. As an additional constraint, participants were required to complete the task without looking directly at the object; instead, they looked at a visual target on the wall in front of them. Eye-tracking glasses were used to record their gaze direction for post-study analysis. This difficult task was designed to mimic a multitasking situation in which vision is directed away from the hand, such as when picking up a drink while watching a film. The supplementary video for this paper presents four sample trials from the study from two viewpoints.

Because of the small diameter of the object and the geometry of the prosthetic hand, there is an optimal grasping location and orientation of the prosthetic hand with respect to the object. Thus, we hypothesized that haptic feedback of contact location could help guide participants to the optimal grasping posture and location. Furthermore, grasping the object with excessive force causes it to slide out of the grasp. The over-grasp controller is designed to help prevent this manipulation error. Finally, slips that occur during the pick-up or set-down phase of the reach-to-pick-and-place task can be prevented by the anti-slip controller. Thus, the envisioned task evaluates both volitional and reflexive components of the sensorimotor-inspired system.

Two stationary bins were used, as shown in Figs.~\ref{fig:Hand} and \ref{fig:setupNoDAQ}. The task began with the object in the start bin ($3.8 \times 3.8 \times 7.6$ cm) and was complete when the object was placed in the end bin ($3.8 \times 3.8 \times 5.1$ cm). The bin centers are 17.5\,cm apart.

\subsubsection{User Study Procedure}
Participants were randomized into two groups to conduct the task using either a standard myoelectric prosthesis (standard condition) or a myoelectric prosthesis with the vibrotactile feedback and tactile reflex control detailed in Sections \ref{Sec:VibroFeedback} and \ref{Sec:AutoControl} (tactile condition). For consistency, the tactile sensors were attached to the prosthesis in both conditions but were not used in the standard condition.

The eye-tracking glasses prevented participants from being able to wear their own prescription glasses during the experiment. They were therefore required to demonstrate an ability to read the largest line on a vision chart from a distance of 3~m. All participants passed this test. Participants then completed a demographics survey with questions regarding their occupation, age, gender, handedness, and experience using haptic devices and myoelectric devices. 

After completing the survey, participants donned the prosthesis by inserting their right hand into the wrist brace and tightening the straps to a comfortable level. The experimenter cleaned the skin over the right wrist flexor and extensor muscle groups before attaching the sEMG electrodes. Next, participants completed the calibration procedure for the sEMG signals as described in Section \ref{Sec:calibration}. If participants were assigned to the tactile condition, a C2 tactor was attached to their right bicep near the elbow.

Participants were then allowed to practice controlling the prosthetic hand using their sEMG signals; wrist flexion causes the hand to close, and wrist extension causes it to open. In order to account for drift in the sEMG signals, which is a well-documented phenomenon of sEMG \cite{Kyranou2018CausesProstheses}, the experimenter first demonstrated the degraded control behavior that occurs as a result of signal drift. She then showed how to re-zero the sEMG signals by reaching a specified position located behind the force plate with the prosthetic hand. Participants were allowed to re-zero their sEMG signals at will, or when prompted by an experimenter observing the sEMG signals during the experiment. 

The experimenter then had the participant don the eye-tracking glasses. To calibrate the glasses, participants held out a Tobii eye-tracking glasses calibration card at arm's length, and calibration was completed using iMotions. 

After setup, participants went through a training session in which the experimenter coached them through the best strategy for completing the task. Participants were then allowed to successfully complete the task while observing the hand and the object. After two successful completions, participants were given five minutes of practice time to try doing the task while looking at the wall target, which greatly increases the difficulty. This five-minute-long practice time without direct vision ended early after two successes. 

\begin{figure}[t]
		\centering
		\vspace{1em}
		\includegraphics[width=\columnwidth]{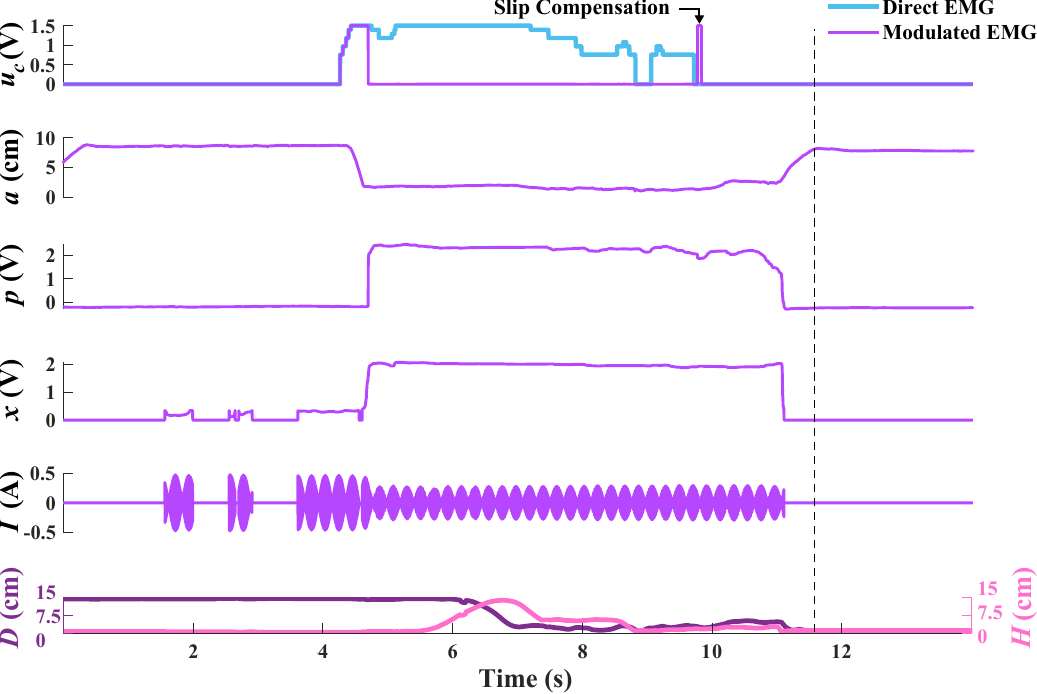}
	    \vspace{-1.75em}
		\caption{Excerpt of time-series traces from a representative participant's trial in the tactile group as they found, grasped, picked up, moved, and set down the object. The dotted line indicates the time point when the participant successfully placed the object into the end bin. The traces shown are the closing command $u_c$, the grip aperture $a$, The pressure sensor signal $p$, the contact location sensor signal $x$, the C2 tactor signal $I$, and the object's displacement $D$ from the end bin and height $H$ above the force plate as measured by the motion-capture cameras. 
 		The participant first attempts to localize the prosthesis hand on the object as shown by the contact-location signal and C2 current traces. Next, the participant activates their EMG, which is modulated once the pressure sensor signal ramps up. 
One fast slip event is also detected from the pressure sensor signal and compensated for.}
		\label{fig:Traces}
		\vspace{-1em}
\end{figure}

Participants then completed twenty trials of reach-to-pick-and-place with the object, always focusing their gaze on the wall target. Each trial started when the experimenter placed the object inside the start bin and was limited to 60 seconds. If the participant was able to complete the task successfully before time was up, the experimenter pressed a button to end the trial. If the participant failed to complete the task in the allotted time, they simply proceeded to the next trial. Sample traces for the motor command, prosthesis aperture, pressure sensor, contact location sensor, C2 tactor, and object location are depicted in Fig.~\ref{fig:Traces} for a representative successful trial from a participant in the tactile condition. 

After the experiment, participants completed a post-experiment survey based on the NASA-TLX questionnaire \cite{Hart1988}. It had a mix of sliding-scale and short-answer questions.

\subsection{Metrics}
Each metric described below was calculated for every trial. The statistical analysis used the mean of each metric across the twenty trials completed by each participant.
\subsubsection{Task Completion Score}
Participants received a score based on the milestones they achieved in the task, as shown in Fig.~\ref{fig:Results}(a). Lifting the object from the start bin merited a score of 1/3 (0.33). If the object was then moved to the vicinity of the end bin, the score increased to 2/3 (0.67). Finally, successfully placing the object in the end bin earned the maximum score of 3/3 (1.00).
\subsubsection{Time Remaining}
The amount of time remaining in the task was also measured. If a participant failed to complete the task, there were 0 seconds remaining. If the participant completed the task just as the allotted 60 seconds ran out, the time remaining for that trial was set to 0.1 seconds. 
\subsubsection{Proportion of Time Spent Looking Away From the Visual Target}
% The proportion of time spent looking away from the wall target was also calculated for each trial using area-of-interest processing of the eye-tracking data and scene video in iMotions.
An automated area-of-interest (AOI) was defined around the visual target for the eye-tracking videos in iMotions. For each trial, this metric was calculated by dividing the amount of time that the participant looked away from the AOI by the total trial time. The resulting proportion shows how much the subject looked away from the wall target; low values indicate compliance, and high values show that the subject may have cheated and looked at the object. 
\subsubsection{Exploration Contact Rate}
We analyzed the contact-location sensor data to count the number of contacts that the prosthesis fingers made before grasping; this value was normalized by dividing it by the trial time to yield a rate. If the trial was unsuccessful, the trial time was 60 seconds. This metric reflects the exploratory procedure used by the participant in order to find the optimal grasping location.
\subsubsection{Fast Slip Rate}
The number of fast slips was calculated using the slip detection algorithm applied to the pressure sensor signal, as described in Section \ref{Sec:AutoControl}. This number was also normalized by dividing by the trial time. Fast slips were likely to occur as a result of over-grasping, or during the pick-up or set-down phases of the task.  

\begin{figure*}[!t]
\centering
	\vspace{1em}
	\includegraphics[width=\textwidth]{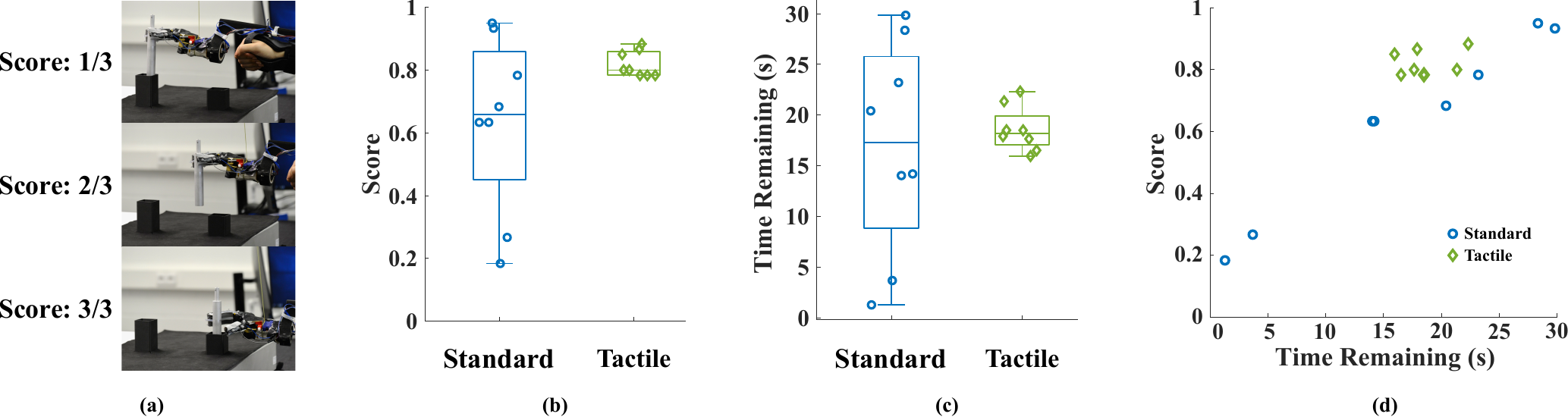}
		\vspace{-1.75em}
		\caption{Task performance. (a) Definition of the scoring metric, which can take values of 0.00 (complete failure), 0.33 (lifted object from start bin), 0.67 (moved object near end bin), and 1.00 (complete success). (b) Average scores for participants in the standard and tactile conditions. (c) Average time remaining in the task for participants in the two conditions. (d) Relationship between score and time remaining for the two conditions.}
		\label{fig:Results}
		\vspace{-1.5em}
\end{figure*}

\subsubsection{Survey}
The post-experiment survey asked participants to separately rate their perceived performance in locating, grasping, lifting, moving, and setting down the object in the end bin. Their average perceived performance was calculated as the average of all five performance ratings. 
% We separately calculated the average perceived performance for lifting, moving, and setting down the object to match our task completion score. 
The survey also asked participants to rate mental effort, physical effort, physical comfort, and time pressure. Next, the survey asked participants to rate their use of visual, auditory, and somatosensory cues. Finally, participants were asked to share any comments on their experience and suggest improvements.

\subsection{Statistical Analysis}
All statistical tests were performed in RStudio (v1.2.1335). 
Levene's test for homogeneity of variance was done for each metric. A MANOVA was performed to jointly assess score and time remaining, which was followed up by linear discriminant analysis (LDA) and individual Welch's t-tests for score and time remaining. Welch's t-test was used to assess the proportion of time spent looking away from the visual target. A robust MANOVA \cite{Munzel2000NonparametricDesigns} was also used to jointly assess score, fast slip rate, and exploratory contact rate. A discriminant analysis using LDA was performed as follow-up, with additional one-way comparisons using Welch's t-tests.
% Welch's T-tests were performed for each of the performance metrics. 

Levene's variance test was also performed for the rating questions. Welch's t-test was used to analyze all of the rating questions in the survey, except time pressure, since this last measure was not normally distributed; Wilcoxon's rank-sum test was performed instead. In addition, the ratings of visual cues were log-transformed before analysis to normalize the distribution. Finally, we report effect sizes using Pearson's correlation coefficient.

\vspace{5pt} 
 
\begin{table*}[!b]
\vspace{-1em}
\caption{Summary of statistics for the survey data. All ratings could range from 0 to 100.}
\vspace{-1em}
\label{survey-table}
\tiny
\centering
\resizebox{\textwidth}{!}{%
\begin{tabular}{l c c c c c c c}%{llllllll} %{l c c c c c c c c c c c c c c c c}
\hline

& \multicolumn{2}{c}{Standard} & \multicolumn{2}{c}{Tactile} &  \multicolumn{3}{c}{Comparison}\\ 
& Mean & SE & Mean & SE & Statistic & $p$ & $r$ \\
Overall Performance & 63.9 & 7.79 & 63.3 & 3.00 & $t$(9.03) = 0.07 & 0.94 & 0.02 \\
% Performance (Lift, move, set down) & 68.8 & 10.1 & 68.8 & 5.46 & $t$(10.8) = 0.00 & 1.00 & 0.00\\
Mental Effort & 56.2  & 9.28 & 62.0 & 4.78  & $t$(10.5) = -0.55 & 0.59 & 0.17\\
Frustration & 47.3  & 7.88 & 58.5 & 5.75  & $t$(12.8) = -1.15 & 0.27 & 0.31\\[-.18em]
Time Pressure & 65.0\footnote[2] & 20.0\footnote[8] & 38.0\footnote[2] & 37.75\footnote[8]  & $W$ = 37.5 & 0.60 & -0.13\\
Physical Effort & 52.0  & 6.32 & 54.0  & 8.62  & $t$(12.8) = -0.19 & 0.85 & 0.05\\
Physical Comfort & 60.0  & 9.50 & 60.1 & 8.80  & $t$(13.9) = -0.01 & 0.99 & 0.003\\
Usage of Auditory Cues & 55.5  & 9.60 & 53.4  & 10.8  & $t$(13.8) = 0.15 & 0.89 & 0.04\\
Usage of Visual Cues & 36.4  & 7.20 & 37.3 & 10.6  & $t$(13.8) = 0.17 & 0.87 & 0.05\\
Usage of Somatosensory Cues & 80.9  & 6.10 & 89.6 & 2.80  & $t$(9.9) = -1.3 & 0.22 & 0.38\\
\hline
\footnote[2]{} Median reported. \footnote[8]{} Interquartile range reported.
\end{tabular}}
\end{table*}

\section{Results}
% \begin{figure*}[!t]
% 		 \centering

% 		\subfloat[]{\includegraphics[width=.13\textwidth]{Figures/scores.pdf}
		
% 		\label{fig:Scores}}
% 		\hfill
%     \subfloat[]{\includegraphics[width=.25\textwidth]{Figures/ScoresTwoCondition.pdf}
		
% 		\label{fig:ScoreBox}}
% 		\hfill
% 		\subfloat[]{\includegraphics[width=.25\textwidth]{Figures/TimeRemTwoConditions.pdf}
	
% 		\label{fig:TimeBox}}
% 		\hfill
% 		\subfloat[]{\includegraphics[width=.25\textwidth]{Figures/scoreVstime.pdf}
	
% 		\label{fig:TimeScore}}
% 		\caption{Task performance. (a) Definition of the scoring metric, which can take values of 0.00 (complete failure), 0.33, 0.67, and 1.00 (complete success). (b) Average scores for participants in the standard and tactile conditions. (c) Average time remaining in the task for participants in the two conditions.}
% \end{figure*}

One participant in the tactile group reported that they could not feel the vibrations output by the C2 tactor on their arm; their data were therefore excluded from the entire analysis, leaving 16 participants. The eye-tracking videos were corrupted for two participants, one of whom was in the standard condition and the other in the tactile condition. Thus, the eye-tracking metrics do not include these two users. For the analyses involving pressure sensor data, two participants in the standard condition were excluded, as the sensor was not functioning during their sessions. One of these participants was also excluded in analyses involving the contact-location sensor's data for the same reason.

% \begin{figure}[t]
% 		\centering
% 		\includegraphics[width=0.75\columnwidth]{Figures/scoreVstime.pdf}
% 		\caption{The average score plotted against the time remaining for each participant, labeled by condition. There is a clear distinction between the standard and tactile groups; indeed, the LDA classification yields 92.8\% accuracy.}
% 		\label{fig:ScoreTime}
% \end{figure}

% \subsection{Performance Metrics}
% There was no significant difference in means for score (), time remaining in task (), and proportion of time spent looking away from the visual target ().
% Visual inspection of the data indicated that the variances of the groups were not equal, which would violate an assumption of the Student's T-test. Indeed,
Fig.~\ref{fig:Results} shows the main task performance results from this study. There was a significant difference in the variance of the scores between groups ($F$(1,14)~=~7.26, $p$~$<$~0.05). There was also a significant difference in variance for time remaining in the task ($F$(1,14)~=~12.86, $p$~$<$~0.01).  %These differences are depicted in Fig. \ref{fig:Results}b and \ref{fig:Results}c. 
There was no significant difference in variance for proportion of time spent looking away from the visual target between the groups (data not plotted, $F$(1,12)~=~0.21), for exploration rate ($F$(1,13)~=~0.40), or for fast slip rate ($F$(1,12)~=~0.37).

The MANOVA indicated a significant effect of condition on score and time remaining ($T$~=~1.9, $F$(2,13)~=~12.56, $p$~$<$~0.001).
A linear-discriminant analysis was used to follow up on the MANOVA (error 6.2\%). 
% which had a classification error of 6.2\%.
The coefficients of the discriminant function revealed differentiation of score ($b$~=~18.16) and time remaining ($b$~= -0.45). The group separation is depicted in Fig. \ref{fig:Results}(d).

Welch's t-tests for score and time remaining were also used as a follow-up to the MANOVA.
The average score was lower in the standard group ($M$~=~0.63, $SE$~=~0.10) than in the tactile group ($M$~=~0.82, $SE$~=~0.01). This difference was not significant ($t$(7.30)~=~-1.85), but it has a large effect size ($r$~$>$~0.5).
The time remaining was lower for the standard group ($M$~=~16.9, $SE$~=~3.74) than the tactile group ($M$~=~18.6, $SE$~=~0.78). This difference was not significant ($t$(7.61)~=~-0.45), and the effect size was small ($r$~=~0.16).
Finally, on average, the proportion of time spent looking away from the visual target was similar in the standard group ($M$~=~0.19, $SE$~=~0.05) and the tactile group ($M$~=~0.18, $SE$~=~0.05). This difference was not significant ($t$(11.94)~=~0.17), and the effect size was small ($r$~=~0.04). \looseness-1

\begin{figure}[t]
		\centering
		\vspace{1em}
		\includegraphics[width=.88\columnwidth]{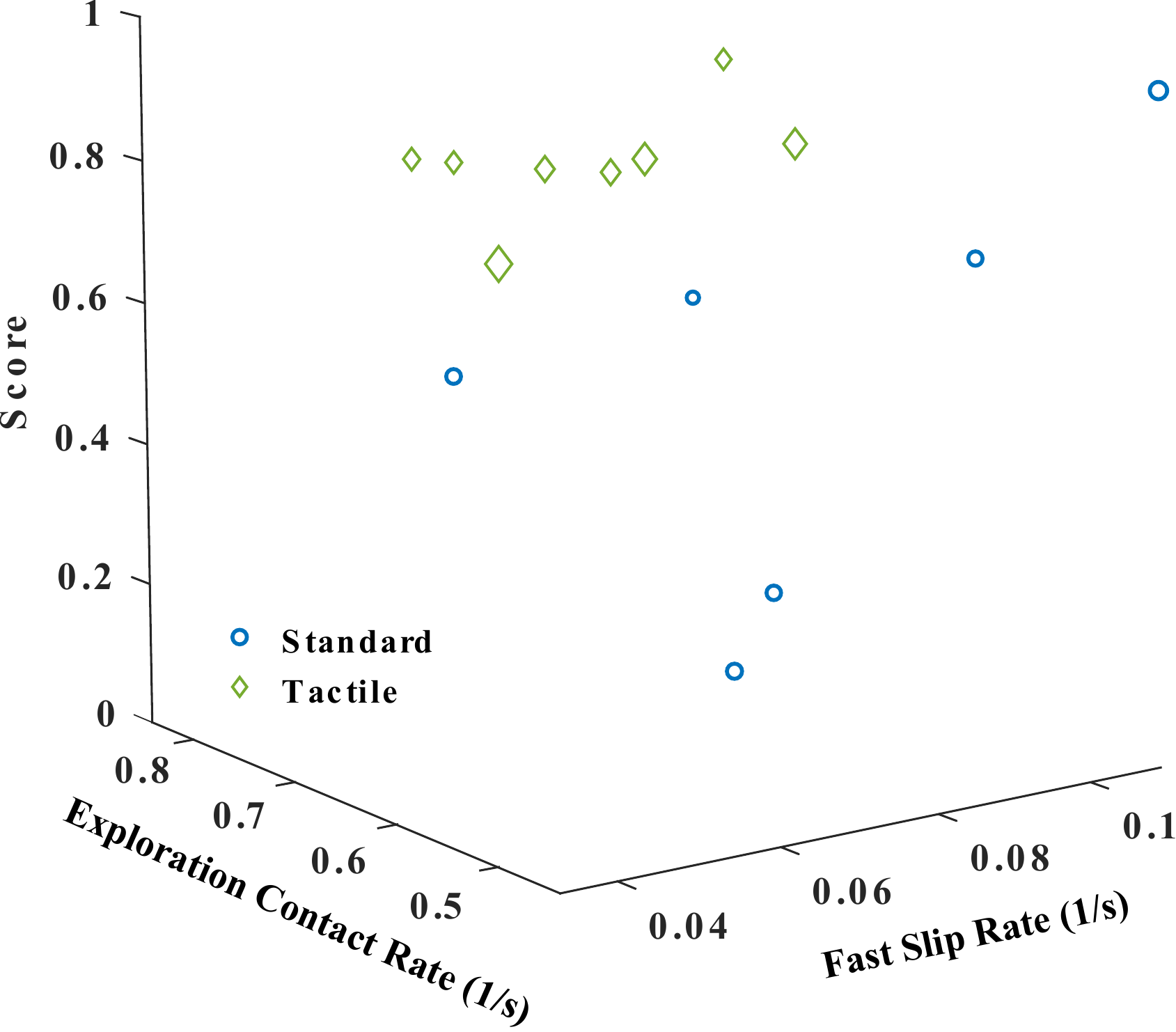}
		\vspace{-.75em}
		\caption{The 3D relationship between exploration contact rate, fast slip rate and task score. The size of the markers varies with depth.}
		\label{fig:ExploreSlipScore}
		\vspace{-1.5em}
\end{figure}

Fig.~\ref{fig:ExploreSlipScore} shows the tactile metrics of the sensorimotor-inspired system plotted against the task performance score. A robust MANOVA showed a significant effect of condition on exploration rate, fast slip rate, and score ($F$~=~2.99, $p$~$<$~0.04). A linear-discriminant analysis was used as a follow-up (error 7.1\%), 
% with classification error 7.1\%, 
in which the coefficients of the discriminant function differentiate exploration rate ($b$~=~8.13), fast slip rate \mbox{($b$~=~-28.6)}, and score ($b$~=~4.78).

Welch's t-tests were also used to follow-up the exploration rate and fast slip rate.
The average fast slip rate was slightly higher in the standard group ($M$~=~0.08, $SE$~=~0.01) than in the tactile group ($M$~=~0.07, $SE$~=~0.007). The difference was not significant ($t$(8.9)~=~0.53) with a small effect size ($r$~=~0.17). The exploration rate in the standard group was lower ($M$~=~0.56, $SE$~=~0.05) than in the tactile group ($M$~=~0.67, $SE$~=~0.05). This difference was not significant ($t$(12.8)~=~-1.81), but the effect size was medium ($r$~=~0.45).

% \subsection{Survey Data}

No differences in variances of the ratings from the survey were found between groups ($F$(1,14) $<$ 3). In addition, no significant differences were found between the conditions for any of the rating questions. See Table~\ref{survey-table} for a summary of these statistics. 

\section{Discussion}

This study investigated the efficacy of haptic feedback of contact location combined with reflex control in an upper-limb myoelectric prosthesis. We compared performance to a standard myoelectric prosthesis in a realistic and difficult reach-to-pick-and-place task in which vision was limited. 
% To date, there has been no research investigating 
% the effect of limited vision on dexterous task performance with a prosthesis. Furthermore, there stood a significant gap in the field regarding 
% the utility of combining haptic feedback autonomous grasp control with haptic feedback in a myoelectric prosthesis. 
When task performance is assessed jointly with score and time remaining, there is a difference between the standard and tactile groups. Namely, although the average time remaining in the task was about the same for the two groups, the participants in the tactile condition more consistently reached a higher task score, indicating high efficiency.
% Our results demonstrate that even with limited vision, prosthesis users were moderately successful in a challenging pick-and-place task without any haptic feedback or aid from automated grasp controllers. 

Consistently grasping the object in the correct place and with the right amount of force was difficult for many participants in the standard condition, as highlighted by high variability in performance. Adding contact-location feedback and reflex control to the prosthesis improved the consistency of task execution, as demonstrated by the significantly lower variability in performance in the tactile condition. It is likely that the tactile reflex control prevented over-grasping, which causes the object to roll out of the grasp, and also prevented slips as the object was being lifted or set down. The utility of the over-grasp reflex control, however, depended entirely on the user's ability to securely and precisely place the fingers of the prosthesis against the object. Here, it is likely that the vibrotactile feedback of contact location helped the user guide the prosthetic hand to the correct grasping location. This is indicated by the result shown in Fig.~\ref{fig:ExploreSlipScore}, where the tactile group's consistent, high scores are supported by their higher exploration rate and lower slip rate. Furthermore, that the exploration and slip rates were not found separately to differ based on group implies that the components of the sensorimotor-inspired system complement each other in different phases of the reach-to-pick-and-place task.

% Due to the constraint on vision and the dimensions of the task object, the task is difficult in two main aspects - finding the right spot to grasp the object, and grasping with the right amount of force. Grasping with too little force would prevent the object from being picked up, and grasping with too much force caused the object to roll away. Furthermore, grasping at any incorrect locations would prevent the object from being picked up. 

These findings parallel those from a separate study investigating the effect of anesthetized fingers on typing when vision of the keyboard and hands was occluded. There, rather than significantly changing mean performance, typing performance variability was significantly impacted \cite{Rabin2004}. In this research, as in our current work, proprioceptive and incidental cues alone seemed to have helped maintain some accuracy, but precision suffers greatly because of the lack of cutaneous feedback and limited vision. The congruence of these results further supports the notion that combining human-in-the-loop haptic feedback of contact location and closed-loop tactile reflexes mimics the natural sensorimotor function of the intact limb.

Interestingly, the two top scorers in the standard condition outperformed everyone in the tactile condition. Participants in the tactile group had to learn how to interpret the haptic feedback in addition to learning how to complete the task. This additional learning step may have caused the difference between the top scorers. Also, one high-scoring participant in the standard group was an aircraft pilot, which requires mastery of difficult visuospatial tasks. The other high-scorer said they could create a non-visual 3D representation of the task area after a few trials. However, their performance was associated with a mental effort rating of 90 out of 100, which is higher than the average score of 56.2 in the standard group. 

One possible explanation as to why there were no differences in means between the two groups is the large variance in the standard condition. Participants undoubtedly have different capabilities, and some adapted to the task better than others. However, participants in the standard group were likely punished more for an inability to adapt than participants in the tactile group. The sensorimotor-inspired system served as a cushion of sorts against poor performance. 
%participants in the sensorimotor group may have relied more on muscle memory and incidental cues than on the vibrotactile feedback. 
%A top-scorer in the standard condition mentioned that as long as they stayed in the same configuration, they were able to do the task moderately well, but had to relearn their approach if they made any postural shifts.
Secondly,
% It is also possible that people came in with different capabilities and adapted to the task better than others. A within-subjects design might have elucidated this idea,  but due to the complex nature of the task, we opted not to put individuals through an even longer session.
based on comments made by a few participants regarding their uncertainty in how to interpret the vibrotactile cues in the tactile condition, performance in this group may have been improved with additional training time with the vibrotactile feedback. Indeed, a study by Stepp {\it et al.} showed that repeated 30-to-45-minute training sessions with both visual and vibrotactile feedback in a virtual pick-and-place task resulted in significant improvement in performance compared to the first session \cite{Stepp2012}. 
% One participant commented that though they could feel the difference between the soft and hard vibrations, the subtler changes once they found the fingertip made it difficult to recognize where the ideal grasping spot was. Another participant suggested providing the feedback only at the optimal grasping spot for the object. Although this most likely would have made the task easier for people in the tactile condition, it may not generalize well to objects of different shapes or sizes. 

We initially chose to use vibrotactile feedback because we could vary the feedback signal continuously, matching the continuous gradient signal of the contact-location sensor. In addition, vibrotactile feedback has been shown to be beneficial in prostheses \cite{Witteveen2014,Thomas2019, Markovic2017} and is low-cost, low-weight, discreet, and easy to implement. However, future investigations should also consider other forms of haptic feedback for contact location, such as distributed pressure. Although using individual pressure tactors would discretize feedback of the continuous contact-location signal, the modality-matched nature of the feedback could improve the results, as less cognitive processing would be required to interpret the feedback \cite{Antfolk2013,Brown2016NonCol}. Another option would be wearable devices that tap, drag across, or squeeze the skin; such feedback was previously shown to outperform vibrotactile cues in motion guidance \cite{Stanley12-TH-Guidance}, though the device would need to be optimized for prosthetics. 
% similar to how modality matched haptic feedback has been previously shown to outperform non-modality matched \cite{}.
% Future work could also investigate alternative ways of providing feedback of contact location, such as discretizing contact location to only a few points.

%the weight of the object is heavier than a normal pen/highlighter, which makes it more distinct during the lift phase. if the object were lighter, the performance in the standard condition may have been worse.
\section{Conclusion}
This paper presented a sensorimotor system concept for upper-limb prosthetics, inspired by the natural functions of human motor control.  We introduced the design of a novel contact-location sensor that is integrated into both the human-in-the-loop haptic feedback and closed-loop tactile reflex control. Our results show that this system standardizes performance in a challenging, dexterous task conducted without direct visual observation. This system would be beneficial for prosthesis users in conditions where vision is limited, such as when lighting conditions are poor, objects are occluded, or when multitasking.
Future work includes testing alternative haptic feedback approaches for the contact location information measured by our sensor. We would also like to validate an improved version of the current system in an amputee population. As a further next step, ways to automatically identify when haptic feedback and automatic grasp control are needed should be investigated. A system that is context-aware and adapts appropriately to the user's needs is necessary in taking upper-limb prosthetics technology to the next level, similar to how intelligent lower-limb prostheses have advanced \cite{Su2019}.

%\addtolength{\textheight}{-5cm}   % This command serves to balance the column lengths

%%%%%%%%%%%%%%%%%%%%%%%%%%%%%%%%%%%%%%%%%%%%%%%%%%%%%%%%%%%%%%%%%%%%%%%%%%%%%%%%

%%%%%%%%%%%%%%%%%%%%%%%%%%%%%%%%%%%%%%%%%%%%%%%%%%%%%%%%%%%%%%%%%%%%%%%%%%%%%%%%

%%%%%%%%%%%%%%%%%%%%%%%%%%%%%%%%%%%%%%%%%%%%%%%%%%%%%%%%%%%%%%%%%%%%%%%%%%%%%%%%

\section*{Acknowledgments}

The authors thank the German Fulbright Commission, the Germanistic Society of America, Mastercard, and the US National Science Foundation Graduate Research Fellowship for funding Neha Thomas. They thank the International Max Planck Research School for Intelligent Systems (IMPRS-IS) for supporting Farimah Fazlollahi. 
Finally, they thank 
% the Stuttgart Max Planck Central Mechanical Workshop for fabricating the object, the MPI-IS Robotics ZWE for 3D-printing parts, and 
Alborz Aghamaleki Sarvestani, Naomi Tashiro, and Fabian Krauthausen for participating in pilot studies.

% Thanks to the German Fulbright Commission, the Germanistic Society of America, Mastercard, and the National Science Foundation Graduate Research Fellowship for funding the first author. 

%%%%%%%%%%%%%%%%%%%%%%%%%%%%%%%%%%%%%%%%%%%%%%%%%%%%%%%%%%%%%%%%%%%%%%%%%%%%%%%%

\bibliography{references_short.bib}
\bibliographystyle{IEEEtranNoURL}

\end{document}